\documentclass[conference]{IEEEtran}
\IEEEoverridecommandlockouts
\usepackage{cite}
\usepackage{amsmath,amssymb,amsfonts}
\usepackage{algorithmic}
\usepackage{graphicx}
\usepackage{tikz}
\usepackage{hyperref}
\usepackage{lipsum}
\usepackage{caption}
\usepackage{siunitx}

\usepackage{subfig}

\usepackage{textcomp}
\usepackage{xcolor}
\def\BibTeX{{\rm B\kern-.05em{\sc i\kern-.025em b}\kern-.08em
    T\kern-.1667em\lower.7ex\hbox{E}\kern-.125emX}}

\newcommand\copyrighttext{%
  \footnotesize \textcopyright 2019 IEEE. Personal use of this material is permitted. Permission from IEEE must be obtained for all other uses, in any current or future media, including reprinting/republishing this material for advertising or promotional purposes, creating new collective works, for resale or redistribution to servers or lists, or reuse of any copyrighted component of this work in other works.}
\newcommand\copyrightnotice{%
\begin{tikzpicture}[remember picture,overlay]
\node[anchor=south,yshift=10pt] at (current page.south) {\fbox{\parbox{\dimexpr\textwidth-\fboxsep-\fboxrule\relax}{\copyrighttext}}};
\end{tikzpicture}%
}

\begin{document}

\title{Effect of Architectures and Training Methods on the Performance of Learned Video Frame Prediction}

\author{M. Akin Yilmaz and A. Murat Tekalp \\
Department of Electrical and Electronics Engineering, Koc University,
Istanbul, Turkey \\
mustafaakinyilmaz@ku.edu.tr, mtekalp@ku.edu.tr
\thanks{This work was supported by TUBITAK project 217E033. A. Murat Tekalp also acknowledges support from Turkish Academy of Sciences (TUBA).}
}
\maketitle

\copyrightnotice

\begin{abstract}
We analyze the performance of feedforward vs. recurrent neural network (RNN) architectures and associated training methods for learned frame prediction. To this effect, we trained a residual fully convolutional neural network (FCNN), a convolutional RNN (CRNN), and a convolutional long short-term memory (CLSTM) network for next frame prediction using the mean square loss. We performed both stateless and stateful training for recurrent networks. Experimental results show that the residual FCNN architecture performs the best in terms of peak signal to noise ratio (PSNR) at the expense of higher training and test (inference) computational complexity. The CRNN can be trained stably and very efficiently using the stateful truncated backpropagation through time procedure, and it requires an order of magnitude less inference runtime to achieve near real-time frame prediction with an acceptable performance.
\end{abstract}


\section{Introduction}
With the advent of deep neural networks, we can now predict future frames in a video with reasonable precision. Frame prediction can be considered as an unsupervised learning problem, since unprocessed (unlabeled) video frames are used as ground-truth in the training process. 

The success of recurrent neural network (RNN) architectures in sequence modeling tasks inspired many to design hybrid networks containing both recurrent and convolutional blocks for video processing. On the other hand, it is also possible to use fully convolutional neural networks (FCNN) for multi-frame video processing by stacking a number of neighboring frames as input to a FCNN. These approaches are analogous to recursive vs. nonrecursive nonlinear filtering.

Most state-of-the-art learned frame prediction (LFP) methods use some form of a recurrent architecture, which are reviewed in Section~\ref{related}. As an alternative, we recently proposed a fully convolutional architecture for LFP with application to predictive video compression~\cite{ssulun_thesis}. However, there are no papers that compare the performance of recurrent and fully convolutional architectures for LFP on a common dataset. To this effect, this paper provides a comparison of various architectures and training methods for LFP in terms of both prediction performance and computational complexity.

Section~\ref{related} reviews related works on video frame prediction. Architectures and training methods for LFP are described in Section~\ref{deeppred}. Experimental evaluation of performance and complexity of different architecture and training methods is presented in Section~\ref{eval}. Section~\ref{conc} concludes the paper.


\section{Related work}
\label{related}
\vspace{-2pt}
Review of recent learned video frame prediction methods can be found in \cite{ssulun_thesis,survey}. We can classify related work on LFP considering their architecture, prediction methodology and loss function used in training. 

In terms of network architecture for LFP, most methods use convolutional LSTM~\cite{srivastava,mathieu,kalchbrenner,pixelcnn,cricri,denton2018stochastic,amersfoort,vondrick,villegaslongterm,wichers2018hierarchical,finn,babaeizadeh2017stochastic,lee2018stochastic}, but there are also some that use fully-convolutional~\cite{ssulun_thesis} networks.

In terms of prediction methodology, we classify related works as \textit{frame reconstruction} methods, which directly predict pixels ~\cite{srivastava,mathieu,kalchbrenner,pixelcnn,cricri,denton2018stochastic} or \textit{frame transformation} methods, which predict transformation parameters, e.g., affine motion parameters, to transform prior frames into future frames~\cite{amersfoort,vondrick,villegaslongterm,wichers2018hierarchical,finn,babaeizadeh2017stochastic,lee2018stochastic}, rather than modeling pixels.

The choice of the loss function depends on whether the evaluation criterion is peak signal to noise ratio (PSNR) or visual detail and sharpness. In the former case, the mean square loss is the only choice. If the predicted images will be viewed by humans, we prefer natural-looking predicted frames without blur. Then, the use of KL-divergence \cite{mathieu}, \cite{vondrick}, \cite{villegaslongterm}, \cite{lee2018stochastic} or  adversarial loss~\cite{babaeizadeh2017stochastic},~\cite{lee2018stochastic},~\cite{denton2018stochastic} is well justified. Other loss functions to emphasize visual detail include $\ell^1$ loss \cite{mathieu}, cross-entropy loss \cite{srivastava,kalchbrenner,cricri}, the gradient loss~\cite{mathieu}, and perceptual loss in a feature space~\cite{villegaslongterm}. 

The novelty of this paper is to compare different network architectures and training methods to directly predict pixels. We use mean square loss since we evaluate prediction performance in terms of PSNR. What we learn in this study are:  \vspace{-10pt}
\begin{itemize}
 \item The FCNN architecture~\cite{ssulun_thesis} yields better prediction performance compared to recurrent architectures at the expense of high complexity at a rate of 1 frame per sec~(fps).
 \item  It is possible to predict 17 fps (near real-time) with reasonable prediction performance and much smaller memory load using RNN architecture and stateful training. 
\end{itemize}


\section{Deep-Learned Video Frame Prediction}
\label{deeppred}  
\vspace{-2pt}
We describe the recurrent and fully convolutional architectures used in our experiments in Sections~\ref{rnn} and~\ref{cnn}, respectively. We used the UCF101 dataset~\cite{ucf101}, converted to grayscale, to train both types of networks. However, preparation of training samples for the recurrent and convolutional networks differ, which is explained under each subsection.

\subsection{Recurrent Neural Network Architectures} 
\label{rnn}
This section describes recurrent neural network architectures and their training for next frame prediction.

\subsubsection{Convolutional RNN (CRNN) and Conv. LSTM (CLSTM)} 
\label{rnnmodel}
CLSTM~is a widely used network architecture for frame prediction~\cite{convlstm}. CRNN is rarely used because of the well-known vanishing/exploding gradients problem while training via backprop-through-time method. In this work, we show that a stateful truncated backprop-through-time procedure, described in Section~\ref{rnntrain}, makes training a CRNN as stable as training a CLSTM, but computationally more efficiently.

\begin{figure}[h]
\centering
	\includegraphics[scale=0.28]{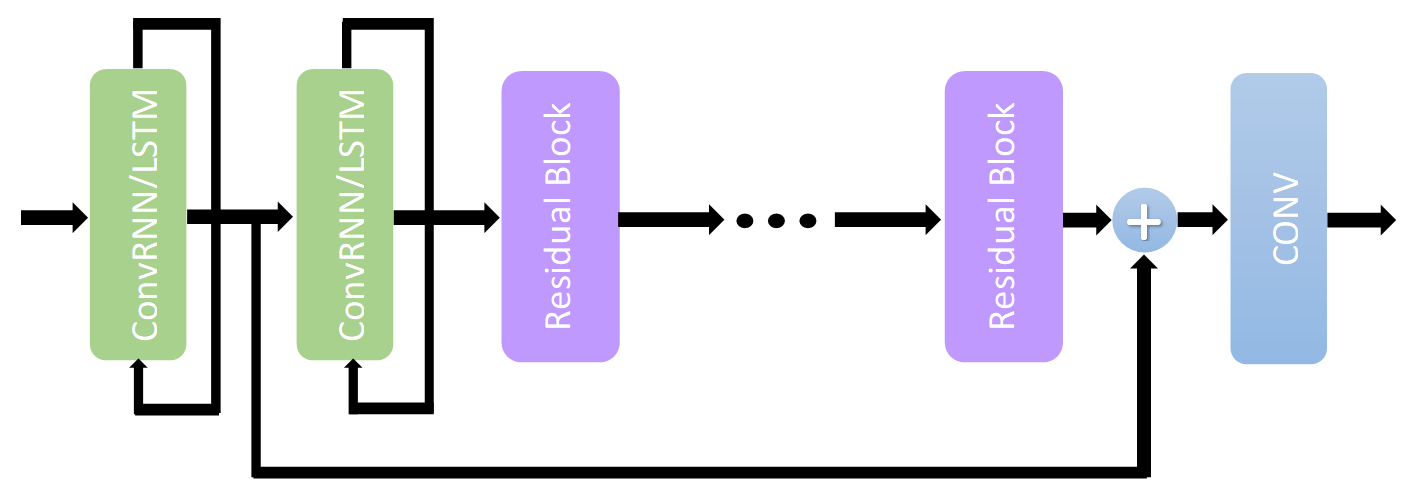} \vspace{-2pt} \\
	(a) \vspace{5pt}\\
	\includegraphics[scale=0.23]{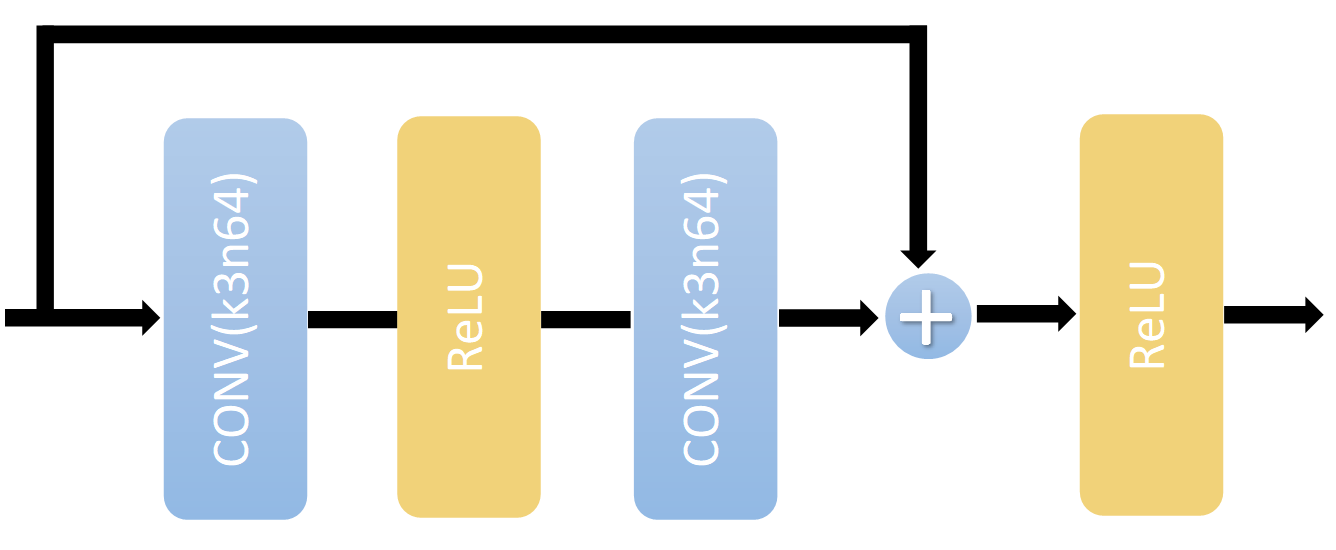} \vspace{-2pt} \\
	(b)
\caption{Block diagram of: a) our CRNN/CLSTM network with two recurrent layers, residual blocks and convolutional output layer, (b) expansion of each purple residual block in (a)}
\label{fig:crnn}
\end{figure}

Our recurrent architecture, with two CRNN/CLSTM layers having depth 64 channels, kernel size 3, and standard activations, is depicted in Fig.~\ref{fig:crnn}. We input grayscale frames one by one to the first CRNN/CLSTM layer to extract temporal information from raw images. Each CRNN/CLSTM layer updates hidden states (and cell states in CLSTM) which are initially set to zero after processing each frame. The hidden state of the second CRNN/CLSTM layer is passed to the first residual block shown in purple. We use 8 residual blocks, with ReLU activation and no residual scaling, all having the same channel depth and kernel size to increase the model capacity. A single convolution layer with linear activation is used as output layer. In all convolution operations, we use stride 1 and padding 1 to preserve dimensions of the input frame. Our networks can process frames with arbitrary height and width.

\subsubsection{Stateless and Stateful Training} 
\label{rnntrain}
We first present the common steps in both stateless and stateful training procedures. We use the well known truncated backpropagation through time (TBPTT) algorithm~\cite{pascanu2013} together with Adam optimizer \cite{adam} with a constant learning rate of 1e-5 and mini-batch size of 4. We randomly pick a set of videos which are longer than 96 frames from the dataset. We normalize input and target patch pixel values to be between -1 and 1 to boost the speed of training convergence. The output pixel values are first truncated to between -1 and 1, then scaled between 0 and 255 for inference. 

For stateless training, we chose to truncate backpropagation after 8 frames. For each minibatch, we select 4 videos at random. For each video, we select a start frame at random. Then, we crop a patch sequence with dimensions $184 \times 184$ pixels with duration of 8 frames, where the position of the crop in the start frame is selected randomly. Both our input and target sequences have 8 consecutive frames; however, the target sequence is 1 frame delayed with respect to the input sequence. At the beginning of each forward pass, we initialize the recurrent state as zeros and recursively update the state vector as each frame is processed. At the end of processing each sequence, we calculate the MSE only between the last frame of the target sequence and its prediction. We backpropagate through all 8 frames and then use Adam optimizer to update the model parameters. In order to backpropagate the loss through 8 time steps, we need to store the gradient vector at each time step. As a consequence, more memory load is needed for training as we use a larger truncation size.

In stateful training of recurrent models, we again select 4 videos at random with a random starting frame. However, we now crop a patch sequence with dimensions $184 \times 184$ pixels and duration 96 frames from each video. Again, the target sequence is 1 frame delayed with respect to the input sequence. In this case, we perform the forward pass for one frame at a time, calculate the loss using each target frame, then backpropagate the loss through time for just a single step. Like all recurrent models, we initialize the state vector to zeros at the start of each minibatch, however, we do not reset states after each forward and backward pass, but we re-initialize states as constant with the previous state vectors in order not to lose temporal information. This procedure requires very low memory consumption during training because we do not need different allocations on memory for gradient vectors at different time steps.
Unlike stateless training, this procedure is not restricted to learn temporal information from a fixed duration sequence limited by the truncation size. Re-initialization of states with the previous state vector after each forward and backward pass enables utilization of the entire sequence for learning temporal information with low memory consumption. Furthermore, stateful training with truncation size 1 brings us 8 times more parameter updates compared to the stateless training procedure for the same number of processed frames.

\subsection{Fully Convolutional Neural Network (FCNN)} 
\label{cnn}

\subsubsection{Model Architecture}
\label{cnnmodel}
The FCNN model is inspired by the enhanced deep super-resolution network (EDSR) \cite{edsr}, the first place entry in the NTIRE~2017 Challenge on Single Image Super-Resolution (SISR) \cite{ntire}. 
The~EDSR network takes a single low resolution colored image as input and produces a higher resolution color image. In order to generate a higher resolution output image, it uses an upsampling block, called the pixel shuffler layer \cite{pixelshuffler}. 

\begin{figure}[t]
\centering
	\includegraphics[scale=0.26]{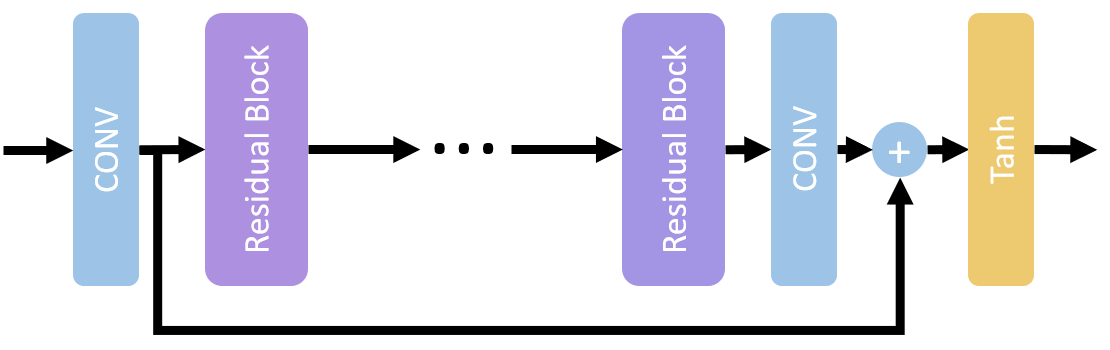} \vspace{-4pt} \\
	(a) \vspace{8pt}\\
	\includegraphics[scale=0.22]{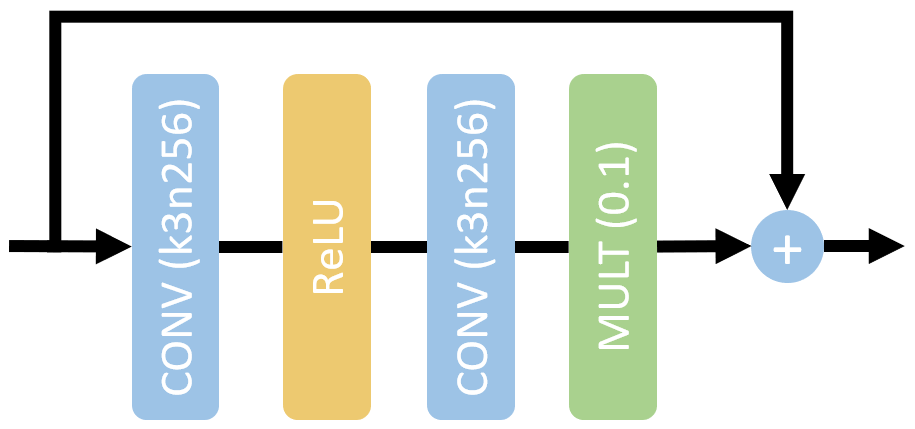} \vspace{-2pt} \\
	(b)
\caption{Block diagram of the CNN: (a) the~EDSR network with modified input and output layers, (b) each residual block of the EDSR network shown in purple in (a).}
\vspace{-10pt}
\label{fig:resnet}
\end{figure}

In contrast, the FCNN takes $K$ past grayscale frames~as input and outputs a single grayscale frame, whose height~and width are the same as those of input frames. To this effect, we modified both the input and output layers of the EDSR network as depicted in Figure~\ref{fig:resnet}~\cite{ssulun_thesis}.  Since the~input and output frames are the same size, we don't need an upscaling layer at the output. There are 32 residual blocks, all with convolution kernel size 3 and channel depth 256. After each residual block, residual scaling by 0.1 is applied \cite{residualscaling}. At the output layer, hyperbolic tangent is used as the nonlinearity, ensuring output values are between -1 and 1. During inference, the outputs are scaled between 0 and 255 to generate predicted video frames. The~model is fully convolutional; hence, it can process inputs with arbitrary height and width. 

We observed that using larger networks consistently provide lower mean-square error similar to that observed in the SISR experiments. This is because the task of image/video generation is so complex that overfitting is rarely observed.

\subsubsection{Training} 
\label{cnntrain}

Our training dataset consists of two~million gray-level (Y-channel) patch sequences, extracted from the UCF101 \cite{ucf101} dataset, where each patch sequence is $48 \times 48$ pixels with duration 9 frames. For each patch sequence, we predict the 9th frame given the first 8, and use the 9th frame as the ground-truth. While extracting patch sequences, we selected the video, the starting frame, and patch position on the start frame randomly. A patch sequence is accepted if it contains sufficient motion, i.e., the sum square difference between successive patches exceeds a motion threshold. Patch sequences that do not satisfy this condition are accepted with probability 0.05. We pick a set of patch sequences randomly from the entire patch sequence dataset to construct minibatches at training time, where one training step corresponds to processing one minibatch. Note that although our training is patch based, we predict full frames at test time.

The model takes 8 frames as input and predicts the 9th frame; we compute the~mean square loss only over the 9th patch. We used Adam optimizer \cite{adam} with an initial learning rate of 1e-4 and a batch size of 32. If the training loss did not decrease for 6000 steps, we halved the learning rate. We trained our model for 400,000 steps (iterations) for fair comparison.


\section{Experimental Evaluation}
\label{eval}
This section presents experimental evaluation of the performances of the stateless and stateful training of both the CRNN and CLSTM networks, denoted by CRNN-stateless, CRNN-stateful, CLSTM-stateless, CLSTM-stateful, respectively, and the fully convolutional network (FCNN). We also evaluate training complexity, model complexity, and test runtimes. 

\begin{figure}[b]
\vspace{-8pt}
\centering
	\includegraphics[scale=0.8]{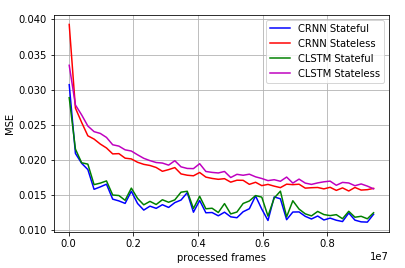} \vspace{-4pt} \\
\caption{Loss vs. number of processed frames.}
\label{fig:test_loss}
\end{figure}

\subsection{Test Set}
\label{datatest}
While we used the UCF101 dataset for training, we used 8 MPEG sequences, Coastguard, Container, Football, Foreman, Garden, Hall Monitor, Mobile, and Tennis, as our test set. Although we used cropped sequences of patches for training, we predict full frames in all methods at test time.

\begin{figure*}
\centering
\subfloat      {
\includegraphics[width=.80\linewidth]{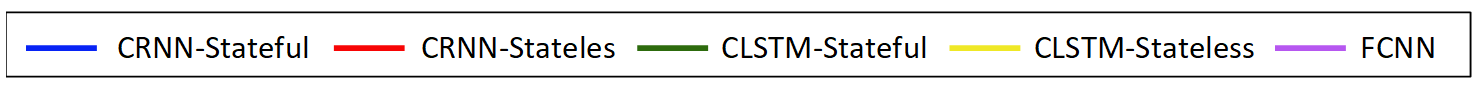}
} \vspace{-10pt} \\
\subfloat      {
\includegraphics[width=.266\linewidth]{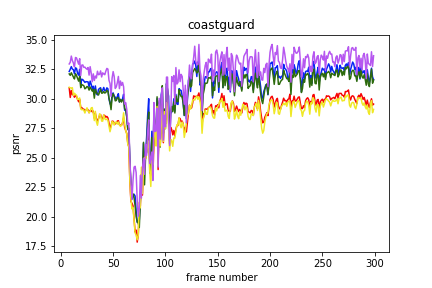}
} \hspace{-23,5pt}
\subfloat     {
\includegraphics[width=.266\linewidth]{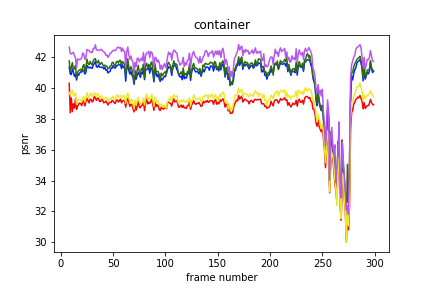}
} \hspace{-23,5pt}
\subfloat     {
\includegraphics[width=.266\linewidth]{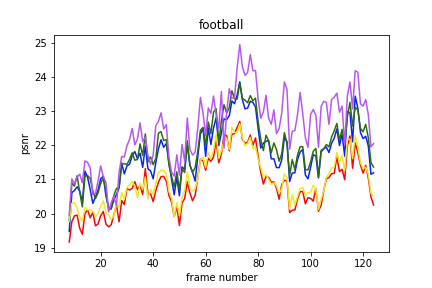}
} \hspace{-23,3pt} 
\subfloat     {
\includegraphics[width=.266\linewidth]{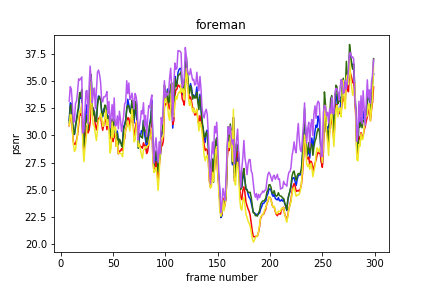}
} \vspace{-5pt} \\
\subfloat     {
\includegraphics[width=.266\linewidth]{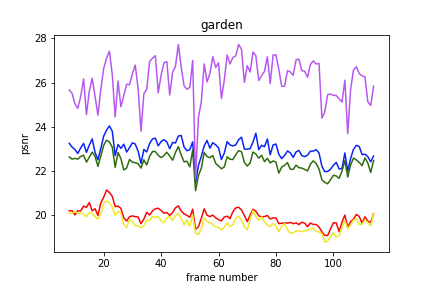}
} \hspace{-23,5pt}
\subfloat     {
\includegraphics[width=.266\linewidth]{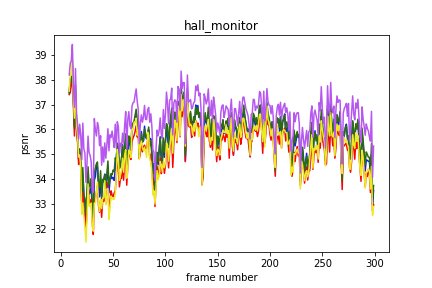}
}\hspace{-23,5pt}
\subfloat     {
\includegraphics[width=.266\linewidth]{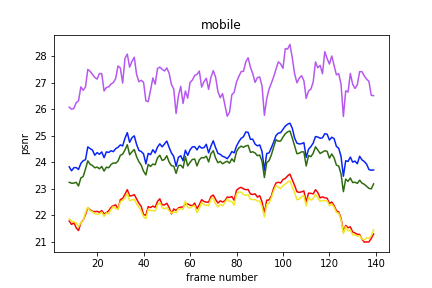}
} \hspace{-23,3pt}
\subfloat     {
\includegraphics[width=.266\linewidth]{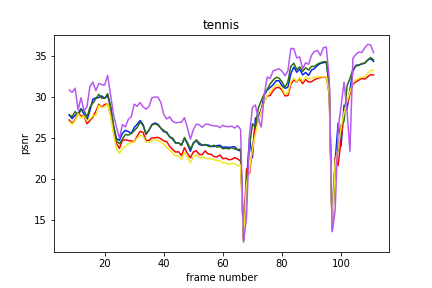}
} \vspace{2pt}\\

\caption{Comparison of prediction PSNR vs. frame number for different models and training methods on MPEG test videos.}
\label{fig:prediction}
\end{figure*}

\subsection{Training: Convergence and Memory Load Comparison}
\label{res:train_load}
The loss as a function of number of frames processed is depicted in Fig.~\ref{fig:test_loss}. We observe that for both the CRNN and CLSTM networks, our stateful training procedure converges much faster than stateless training since we perform parameter updates after processing each frame as opposed to once every 8 frames in stateless training. On the other hand, Fig.~\ref{fig:test_loss} shows that stateful training yields a somewhat noisier convergence compared to the stateless procedure because we perform parameter updates after processing each frame. The rate of decay of the loss curves suggests that the stateless training curves (purple and red) may reach lower loss value if we continue to process more minibatches, which shows that stateless training requires more computation to reach a similar loss to that of stateful training. We see that the loss for CRNN decays faster as it has less parameters compared to the CLSTM model.

Furthermore, stateful training also requires lower GPU memory usage. The GPU memory usage of stateful and stateless training of CRNN and CLSTM models are shown in Table~\ref{table:gpu}. We note that training the FCNN model requires 12 GBytes, which is the entire GPU memory. Low memory consumption makes it possible to train bigger models that cannot fit into the GPU in case of stateless training. It also allow us to use larger batch sizes for more robust and less noisy training.

\begin{table}[!h]
\centering
\caption{GPU memory load (MB) during training}
\begin{tabular}{c|c|c}
             & Stateful &   Stateless   \\ \hline
CRNN          & 1783   &    3353 \\
CLSTM         & 2661       &    9217   \\
\end{tabular}
\label{table:gpu}
\end{table}

\subsection{Comparison of Prediction Performance}
\label{pred}
In this section, we present quantitative evaluation of prediction performance. To this effect, we plot the PSNR of frames predicted by the CRNN-stateless, CRNN-stateful, CLSTM-stateless, CLSTM-stateful, and fully convolutional models vs. frame number for each test video in Fig.~\ref{fig:prediction}. 

Results in Figure~\ref{fig:prediction} shows that FCNN gives the highest PSNR on all test videos because it has a larger number of parameters as shown in Table~\ref{table:complexity}. However, statefully trained CRNN and CLSTM models provide competitive results in all videos, except for garden and mobile, which have camera motion, using a much lower number of parameters. This is a clear indication that more model parameters are needed to learn more complex motions. 

The stateless trained CRNN and CLSTM models perform the poorest, mainly because they are not sufficiently well trained as quickly as the models obtained by the stateful training procedure, where we trained all models by processing the same number of training frames as shown in Fig.~\ref{fig:test_loss}.

\subsection{Comparison of Model Complexity and Test Runtime}
\label{complexity}

CRNN and CLSTM models have much less complexity in terms of both number of parameters and test runtime compared to FCNN as shown in Table~\ref{table:complexity}. Due to its non-recurrent nature, FCNN processes the same frames multiple times. Combination of its heavy structure and the need for 8 frames to produce a single frame, FCNN can only generate 1 frame per second (fps) on a single NVIDIA GeForce GTX 1080Ti GPU on an HP Server with Intel Xeon Gold CPU @ 2.30GHz with 24 cores. On the other hand, the smaller size of CRNN and CLSTM models make it possible to generate 17 fps on the same hardware. 
\vspace{5pt}

\begin{table}[!h]
\centering
\caption{Model Complexity and Test Runtime}
\begin{tabular}{c|cccc}
             & CRNN & CLSTM &  FCNN 
\\ \hline
\# of parameters   & 702,849   & 1,037,121  &  38,376,193
\\
Runtime (fps)     & 17   & 17    & 1
\\
\end{tabular}
\label{table:complexity}
\end{table}


\section{Discussion and Conclusions}
\label{conc}

The performance of learned next frame prediction is surprisingly good using both FCNN and recurrent networks. Compared to FCNN, CRNN and CLSTM models have much less complexity. However, when trained by using the stateful learning they are able to compete with FCNN in terms of PSNR with a much lower runtime. These attributes make CRNN or CLSTM a suitable candidate for frame prediction.

A more detailed inspection of the results reveal the following: 
1) The PSNR performance of the FCNN network is the best on all test videos. 
2) Results obtained by stateful training are superior to that of stateless training for both CRNN and CLSTM on all test videos, since stateful training converges faster.
3) Stateful training procedure requires less memory footprint since we do not need to store gradient vectors for other frames.

A potential application of learned video frame prediction is predictive video compression as shown in~\cite{ssulun_thesis}. It is observed that training using mean square loss may produce blurry predicted images; however, the proposed architectures and training procedures can be used by other loss functions depending on the application.

\clearpage

\bibliography{references}
\bibliographystyle{IEEEtran}

\end{document}